\documentclass[10pt,twocolumn,letterpaper]{article}

\usepackage{cvpr}
\usepackage{times}
\usepackage{epsfig}
\usepackage{graphicx}
\usepackage{amsmath}
\usepackage{amssymb}
\usepackage{authblk}
\usepackage{algorithm}
\usepackage{algorithmic}

\usepackage{caption}
\usepackage[pagebackref=true,breaklinks=true,letterpaper=true,colorlinks,bookmarks=false]{hyperref}
\usepackage{cleveref}

\cvprfinalcopy 

\def\cvprPaperID{9562} 

\begin{document}

\title{RealMix: Towards Realistic Semi-Supervised Deep Learning Algorithms}
\author[1,2]{Varun Nair}
\author[2]{Javier Fuentes Alonso}
\author[2]{Tony Beltramelli}
\affil[1]{Duke University, Durham, NC, USA}
\affil[2]{Uizard Technologies, Copenhagen, Denmark}
\affil[ ]{\tt\small {varun.nair1@duke.edu, javier@uizard.io, tony@uizard.io}}
\maketitle

\begin{abstract}
   Semi-Supervised Learning (SSL) algorithms have shown great potential in training regimes when access to labeled data is scarce but access to unlabeled data is plentiful. However, our experiments illustrate several shortcomings that prior SSL algorithms suffer from. In particular, poor performance when unlabeled and labeled data distributions differ. To address these observations, we develop RealMix, which achieves state-of-the-art results on standard benchmark datasets across different labeled and unlabeled set sizes while overcoming the aforementioned challenges. Notably, RealMix achieves an error rate of 9.79\% on CIFAR10 with 250 labels, and is the only SSL method tested able to surpass baseline performance when there is significant mismatch in the labeled and unlabeled data distributions. RealMix demonstrates how SSL can be used in real world situations with limited access to both data and compute and guides further research in SSL with practical applicability in mind.
\end{abstract}

\section{Introduction}
\label{introduction}

Recent progress in deep learning has largely been driven by the development of specialized hardware and the abundance of large, labeled datasets. While applicable in learning tasks when data is widely and cheaply available, these techniques are impractical to solve real world problems where collecting data is both time-consuming and expensive. Typical examples of such problems include diagnosis from medical imaging and robotic perception problems.

\begin{figure}[t!]
    \includegraphics[width=1.\linewidth]{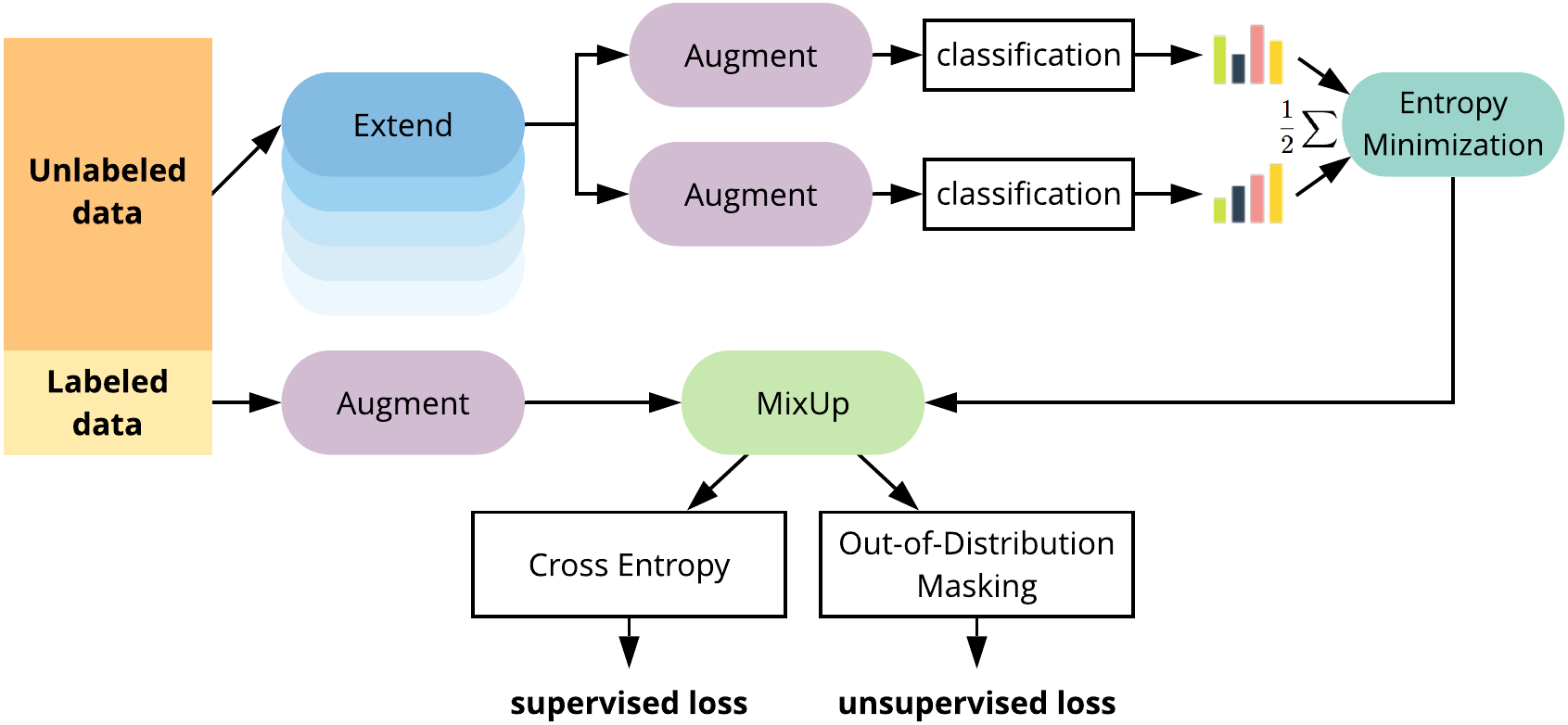}
    \caption{A high-level illustration overview of RealMix, a novel semi-supervised learning technique improving classification performance when there is a  significant shift between the distributions of the unlabeled and the labeled data.}
    \label{fig:realmix_overview}
\end{figure}

To combat challenges in these domains, Semi-Supervised Learning (SSL) algorithms have emerged as a useful tool~\cite{introsslChapelle}. SSL algorithms seek to learn the underlying structure of data by utilizing large amounts of unlabeled data, which can often be more readily available than labeled data. Recent work in SSL~\cite{mixmatchBerthelot, vatMiyato, meanteacherTarvainen, ictVerma} has progressed using a number of assumptions. First, that model outputs on unlabeled data should be invariant to small perturbations (i.e. consistency training). Second, that encouraging model outputs to be more confident will steer decision boundaries away from high-density regions (i.e. entropy minimization~\cite{entropyminsslGrandvalet}). Finally, that the training data distribution can be extended using linear interpolations of data points (i.e. MixUp~\cite{mixupZhang}).

SSL algorithms are typically evaluated by taking a standard benchmarking dataset (e.g. CIFAR10~\cite{cifar10krizhevsky}, SVHN~\cite{svhnNetzer}) and discarding a significant fraction of the labels. This results in a small labeled dataset and a larger unlabeled dataset that both come from the same distribution. The current state-of-the-art SSL technique MixMatch~\cite{mixmatchBerthelot} is able to recover over 92\% of the test accuracy on CIFAR10 using 200 times fewer labels than the supervised baseline. These advances prompt the following question: Can SSL algorithms sufficiently alleviate the need for labeled data in real-world settings?

Oliver et al.~\cite{realisticevalOliver} argues that the current approach to evaluating SSL algorithms is inadequate and raises several questions about SSL's real-world applicability. In particular, they find that performance of SSL techniques suffer when there is a significant mismatch in the unlabeled and labeled data distributions and that transfer learning can often outperform SSL with labeled data alone. We reevaluate these findings on sections 4.2.2 and 4.2.3, showing that this is no longer true.

These problems have, up until now, been a major drawback on the adoption of SSL techniques in realistic setups. We can define a realistic setup for SSL as one in which a practitioner compares SSL performance with transfer learning using limited labeled samples (given its success in pre-training classifiers~\cite{transferlearningBengio}) and where unlabeled data samples are not guaranteed to come from the labeled data distribution. Our goal is to develop a deep SSL algorithm that unites successful practices in SSL and is viable in realistic setups.

We present RealMix, an SSL algorithm depicted in \cref{fig:realmix_overview} that unites ("mixes") the most successful approaches in SSL to set state-of-the-art results on benchmark datasets while surpassing baseline performance when there is significant mismatch in the unlabeled and labeled datasets. Our contributions can be summarized as follows:
\begin{itemize}
    \item We perform experiments to show that RealMix sets state-of-the-art results on CIFAR10 and SVHN, achieving an error rate of \textbf{9.79\%} on CIFAR10 using 250 labels.
    \item We experimentally demonstrate that RealMix is applicable in real-world settings by showing that when the unlabeled distribution is significantly different from the labeled distribution, we can still improve on the supervised baseline performance. Notably, \textbf{RealMix is the only SSL approach tested that is able to surpass baseline performance when there is significant or complete mismatch in the labeled and unlabeled distributions.}
    \item We demonstrate that RealMix (in addition to MixMatch~\cite{mixmatchBerthelot}) surpasses transfer learning, and that transfer learning is \textbf{complementary} to SSL. We show this experimentally by pre-training a classifier on ILSVRC-2012~\cite{imagenet} and applying RealMix to further reduce the error on CIFAR10 with 250 labels to just \textbf{8.48}\%.
    \item We also perform an ablation study on RealMix to identify the components that lead to its success in realistic scenarios.
    \item We provide our implementation source code as a publicly available repository\footnote{Available at https://github.com/uizard-technologies/realmix} to foster future research.
\end{itemize}

We continue our discussion of RealMix in the next section by detailing successful approaches in SSL, how RealMix unites these approaches, and what new elements are introduced by RealMix to make it work in realistic scenarios. In \cref{experiments:main}, we carry out several experiments with RealMix that lead to state-of-the-art results on benchmark image classification datasets and demonstrate its effectiveness when unlabeled and labeled data distributions mismatch.


\newcommand{\req}{\algorithmicrequire\hspace*{0.5em}}
\newcommand{\tab}{\hspace*{3mm}}
\newcommand{\tabb}{\tab\tab}

\begin{algorithm*}[t!]
\caption{Pseudocode for RealMix Algorithm}
\label{alg:realmix}
\begin{algorithmic}[1]
\STATE \req $f_\theta (x)$: deep neural network with trainable parameters $\theta$\\
\STATE \req $X_l, Y_l$: set of labeled data points
\STATE \req $X_u$: set of unlabeled data points
\STATE \req $Extend(x)$: stochastic data augmentation function for unlabeled data
\STATE \req $Augment(x)$: stochastic data augmentation function for consistency training
\STATE \req $MixUp_{\alpha}(A, B), \alpha$: MixUp function and Beta distribution parameter
\STATE \req $TSA(\mathcal{L}_{sup}), schedule$: Training signal annealing function for supervised loss and annealing schedule.
\STATE \req $OODMask_{\gamma}(\mathcal{L}_{sup}), \gamma$ Out-of-distribution masking function for unsupervised loss and masking parameter.
\STATE $\hat{X}_u = Extend(X_u)$
\FOR {$t$ in $1, \dots, num\_epochs$ }
\FOR {$b$ in $1, \dots, num\_batches$}
\STATE $\hat{x}_{l, b} = Augment(x_{l, b})$
\STATE $\hat{\hat{x}}_{u, b}, y_{u, b} = generateTargets(\hat{x}_{u, b})$ 
\ENDFOR
\STATE $\hat{X}_{l+u} = \hat{X}_{l} + \hat{\hat{X}}_{u}$
\STATE $Y_{l+u} = Y_{l} + Y_{u}$
\STATE $X'_l, Y'_l = MixUp_{\alpha}((\hat{X}_{l}, Y_{l}), (\hat{X}_{l+u}, Y_{l+u}))$\label{line:mixuplabeled}\\
\STATE $X'_u, Y'_u = MixUp_{\alpha}((\hat{\hat{X}}_{u}, Y_{u}), (\hat{X}_{l+u}, Y_{l+u}))$\label{line:mixupunlabeled}\\
\STATE $\mathcal{L}_{sup} = CrossEntropy(f_\theta(X'_l), Y'_l)$
\IF{$TSA$}
\STATE $\mathcal{L}_{sup} = TSA(\mathcal{L}_{sup}, schedule)$
\ENDIF
\STATE $\mathcal{L}_{unsup} = MSE(f_\theta(X'_u), Y'_u)$
\STATE $\mathcal{L}_{unsup} = OODMask_{\gamma}(\mathcal{L}_{unsup})$
\STATE $\mathcal{L} = \mathcal{L}_{sup} + \lambda\mathcal{L}_{unsup}$
\STATE $\theta' = ExponentialMovingAverage(\theta)$
\STATE perform gradient descent update on $\theta'$ using $\mathcal{L}$
\ENDFOR
\RETURN $\theta'$
\end{algorithmic}
\end{algorithm*}

\section{Related Work}
\label{relatedwork}

While SSL techniques have a rich history~\cite{introsslChapelle, clusterChapelle, sslZhu}, we focus on describing methods that recent deep variants utilize to achieve  state-of-the-art and literature that considers SSL in realistic setups.

\subsection{Consistency Training and Data Augmentation}

Chapelle et al.~\cite{introsslChapelle, clusterChapelle} describe the cluster assumption, in which data samples that belong to the same cluster structure are likely to belong to the same class. Unlabeled data points can then be used to better define the boundaries of these clusters, where the class of each cluster is defined by the labeled data points within. This assumption is also equivalent to the low-density assumption, in which the decision boundaries should lie in low-density regions. Consistency training (also consistency regularization) can be formulated by combining these assumptions into a regularization task: given an unlabeled data point $x_u$, a classifier $f_{\theta}(x)$, and a perturbation $\delta$, then $f_{\theta}(x_u)$ = $f_{\theta}(x_u + \delta)$. In other words, a classifier should be invariant to small perturbations applied on the input, which is typically enforced by an additional loss term.

The choice of perturbation ($\delta$) induced on an unlabeled sample has varied across SSL techniques. The $\Gamma$-Model~\cite{laddernetworksRasmus},  $\Pi$-Model~\cite{temporalensemblingLaine}, and Mean Teacher~\cite{meanteacherTarvainen} perturb unlabeled samples using Gaussian noise and simple data augmentations (e.g. random translation and horizontal flips), while VAT~\cite{vatMiyato} applies noise that adversarially affects classifier outputs. UDA~\cite{udaXie} applies a more diverse range of augmentations, and ICT~\cite{ictVerma} and MixMatch~\cite{mixmatchBerthelot} both use MixUp ~\cite{mixupZhang} to train an SSL classifier to output consistent predictions on linear interpolations of data points. 

RealMix performs consistency training by applying MixUp, horizontal flips, and random translation on labeled and unlabeled samples (detailed in \cref{mixup}) We also extend our unlabeled sample distribution by creating several augmented copies (augmented using CutOut~\cite{cutoutDevries}).

\subsection{Entropy Minimization}

Entropy Minimization (EM)~\cite{entropyminsslGrandvalet} has been applied in SSL to encourage high-confidence classifier outputs. This approach is also inspired by the low-density assumption, as a classifier with a decision boundary passing through high-density regions would make low-confidence predictions on a number of samples. VAT~\cite{vatMiyato} incorporates EM as a loss term to further improve results, and MixMatch~\cite{mixmatchBerthelot} and UDA\cite{udaXie} apply EM by sharpening the targets of unlabeled samples. We also apply EM through a sharpening function (described in \cref{entropyminimization}), as we find it to work well experimentally.

\subsection{SSL in Realistic Contexts}

Oliver et al.~\cite{realisticevalOliver} described a number of pitfalls of current SSL algorithms and provided recommendations to practitioners for when SSL may be appropriate. We do not investigate all of their findings, but instead focus on those most pertinent to our work. Specifically, these include that SSL is most likely applicable if: 
 \begin{itemize}
     \item Transfer learning from similar domains using labeled datasets is not feasible. 
     \item The labeled and unlabeled data samples are drawn from the same distribution.
 \end{itemize}

With the two above-mentioned points in mind, we find that: 
 \begin{itemize}
     \item RealMix surpasses performance compared to transfer learning and fine-tuning even when transfer learning from similar domains is feasible, including when the target and transfer domains share classes. We show this experimentally in \cref{transferleanring}.
     \item RealMix is capable of surpassing baseline performance even when upwards of 75\% of the unlabeled data comes from a different distribution than the labeled data. We accomplish this using out-of-distribution masking, which prevents our classifier from learning on examples that are out-of-distribution. This is detailed further in \cref{ofdmasking}.
 \end{itemize}
 

\begin{algorithm*}
\caption{Pseudocode for generating targets}
\label{alg:generateTargets}
\begin{algorithmic}[1]
\STATE \req $Sharpen(d, t)$: entropy minimization function
\STATE \req $\hat{x}_{u, b}$: batch of unlabeled samples
\STATE $\hat{\hat{x}}_{u, b, aug1} = Augment(\hat{x}_{u,b})$
\STATE $\hat{\hat{x}}_{u, b, aug2} = Augment(\hat{x}_{u,b})$
\STATE $y_{u, b,aug1} = f_\theta(\hat{\hat{x}}_{u, b, aug1})$
\STATE $y_{u, b,aug2} = f_\theta(\hat{\hat{x}}_{u, b, aug2})$
\STATE $\bar{y}_{u, b} = \frac12 (y_{u, b, aug1} + y_{u, b, aug2})$
\STATE $\bar{y}_{u, b} = Sharpen(\bar{y}_{u, b}, 0.5)$ \label{line:sharpening} \\
\RETURN $\bar{y}_{u, b}$
\end{algorithmic}

\end{algorithm*}

\section{RealMix}

As discussed in \cref{introduction} and \ref{relatedwork}, RealMix unites the most successful approaches in SSL and adapts them to work in realistic contexts. An overview for RealMix is presented in \cref{fig:realmix_overview} and \cref{alg:realmix}.

Formally, given labeled samples $(X_l, Y_l)$, unlabeled samples $X_u$, MixUp beta distribution parameter $\alpha$, out-of-distribution masking parameter $\gamma$, and consistency training (unlabeled loss) weight $\lambda$, we can obtain a classifier $f_\theta$ that minimizes \cref{eq:loss}: 
\begin{equation}
    f_\theta = RealMix(X_l, Y_l, X_u, \alpha, \gamma, \lambda)
\end{equation}
\begin{equation} \label{eq:loss}
    \mathcal{L} = \mathcal{L}_{sup} + \lambda \mathcal{L}_{unsup}
\end{equation}
where $\mathcal{L}_{sup}$ is the standard cross-entropy loss on labeled samples, $\lambda$ is the consistency training (unlabeled loss) weight, and $\mathcal{L}_{unsup}$ is computed using MSE and out-of-distribution masking (see \cref{experiments:ood}) on targets of unlabeled samples. The generation of unlabeled targets is presented in \cref{alg:generateTargets} and we discuss hyperparameters $\alpha$ and $\gamma$ in the following subsections.

\subsection{Data Augmentation}
\label{dataaugmentation}

Following UDA~\cite{udaXie}, we first extend our unlabeled set $X_u$ by applying 50 rounds of augmentations to $\hat{X}_u$ using $Extend(x)$, where $Extend(x)$ can include cropping, flipping, or stronger augmentations such as CutOut~\cite{cutoutDevries}. By using several augmented copies of unlabeled data, we provide our classifier with a wide range of perturbations that give more inductive biases about the data distribution.

As a part of consistency training, we compute targets $\bar{y}_{u, b}$ for each unlabeled batch $\hat{x}_{u, b}$ by averaging the classifier's predicted distribution over two additional augmentations created by a separate augmentation function, denoted $Augment(x)$ (as shown in \cref{alg:generateTargets}). We settled on two augmentations as additional augmentations significantly increased training time without significantly improving results. Note that $Extend(x)$ produces many copies of unlabeled data, whereas $Augment(x)$ produces just 1 copy for use in generating targets.

\subsection{Entropy Minimization}
\label{entropyminimization}

MixMatch~\cite{mixmatchBerthelot} and UDA~\cite{udaXie} both implement entropy minimization through a sharpening function, which we also find to be helpful. By applying this function (line \ref{line:sharpening} of \cref{alg:generateTargets}) on the unlabeled targets $\bar{y}_{u, b}$, we encourage our classifier to produce low entropy predictions on unlabeled data. That is, for each class $c_i \in C$:
\begin{equation}
    Sharpen(p, t)_i := \frac{p_i^{\frac{1}{T}}}{\sum_{k = 1}^{|C|}p_k^{\frac{1}{T}}} 
\end{equation}

where $p$ is the average predicted class and $t$ is the temperature of the sharpened distribution. Intuitively, the distribution approaches a one-hot distribution as $t$ goes to 0. We find $t = 0.5$ to be a good value across multiple benchmark datasets and use it in all reported experiments.  

\subsection{MixUp}
\label{mixup}
MixUp was proposed by Zhang et al.~\cite{mixupZhang} as a regularization technique to encourage high-margin decision boundaries and was utilized in SSL by ICT~\cite{ictVerma} and MixMatch ~\cite{mixmatchBerthelot}. Given two samples $(x_1, y_1)$, $(x_2, y_2)$ and Beta distribution parameter $\alpha$, our MixUp function generates a new sample $(x_3, y_3)$ as follows:

\begin{align}
    \phi &\sim Beta(\alpha, \alpha) \\
    \phi' &= max(1-\phi, \phi) \\ \label{eq:mixupmax}
    x_3 &= \phi'x_1 + (1-\phi')x_2 \\
    y_3 &= \phi'y_1 + (1-\phi')y_2
\end{align}

Following data augmentation and the generation of unlabeled sample targets, we apply MixUp separately to both the labeled samples $(\hat{X}_l, Y_l)$ and unlabeled samples $(\hat{\hat{X}}_u, Y_u)$ (see lines \ref{line:mixuplabeled}-\ref{line:mixupunlabeled} of \cref{alg:realmix}). As in MixMatch, the resulting samples $(X'_l, Y'_l)$ and $(X'_u, Y'_u)$ are linear interpolations of samples from both the labeled and unlabeled collections but are weighted to more closely resemble their "original" distribution (\cref{eq:mixupmax}). In other words, $(X'_l, Y'_l)$ are more similar to the original labeled points and $(X'_u, Y'_u)$ are more similar to the original unlabeled points. 

\begin{table}[h]
\centering
\begin{tabular}{|l|rrrrr|}
    \hline
    Method & 0\% & 25\% & 50\% & 75\% & 100\% \\ \hline
    MT \cite{meanteacherTarvainen} & $17.85$ & $17.25$ & $18.95$ & $20.14$ & $20.57$\\
    MM \cite{mixmatchBerthelot} & $16.75$ & $18.14$ & $21.01$ & $21.07$ & $22.08$\\\hline
    \textbf{RealMix} & $\bf{16.41}$ & $\bf{16.60}$ & $\bf{16.51}$ & $\bf{16.99}$ & $\bf{17.62}$\\ \hline
\end{tabular}
\caption{Results comparing error of RealMix to other SSL methods on the distribution mismatch experiment. 0\% mismatch serves as the baseline in which the labeled and unlabeled data are drawn from the same distribution. While other methods steadily increase in error as amount of mismatch increases, RealMix is surprisingly able to surpass baseline performance when there is over 75\% mismatch.}
\label{table:oodresults}
\end{table}

\subsection{Out-of-Distribution Masking}
\label{ofdmasking}
\begin{figure}[t!]
    \includegraphics[width=1.\linewidth]{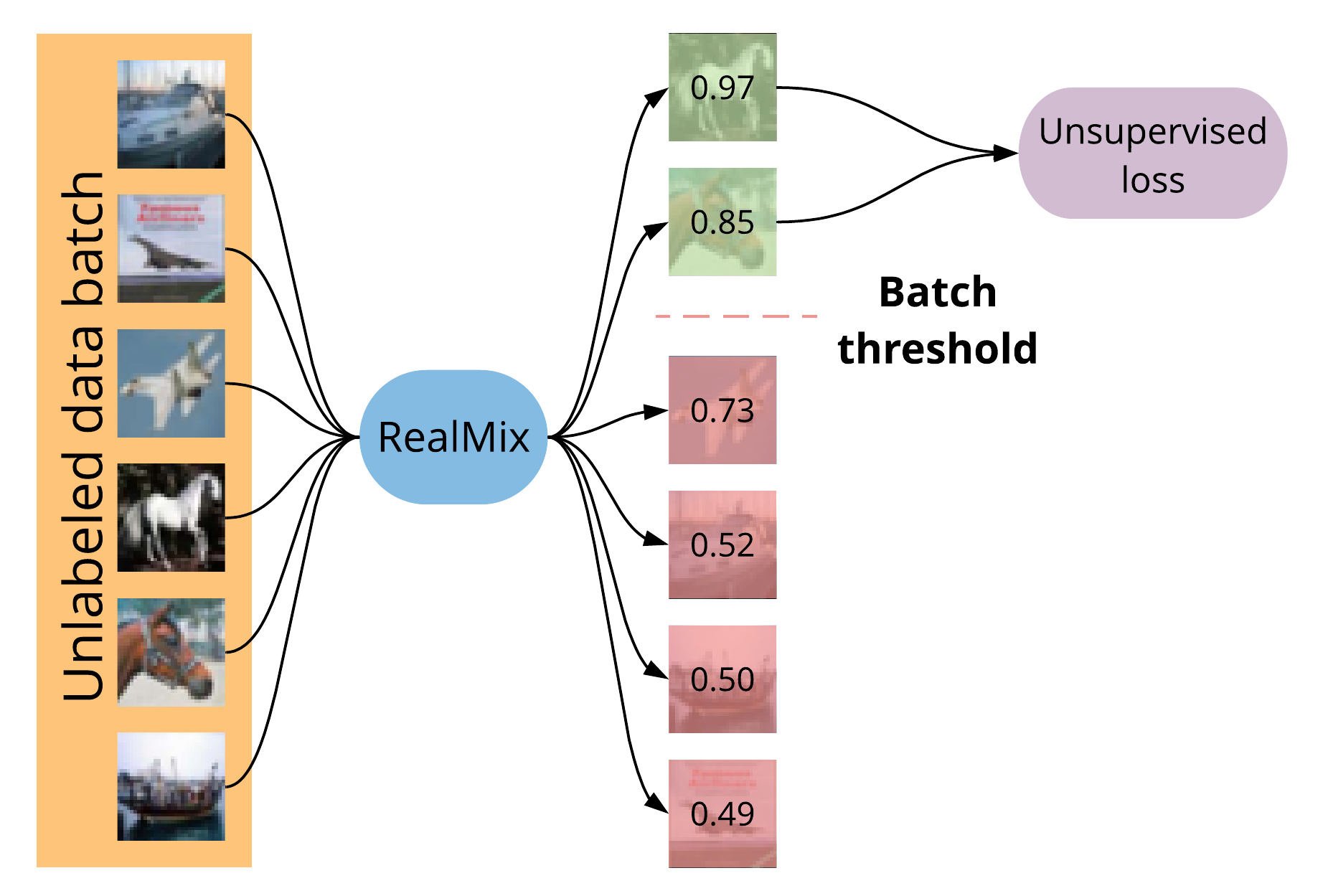}
    \caption{Illustration of the out-of-distribution masking process. RealMix produces both the confidence on each of the images and the threshold that should be applied to them based on $\gamma$ and the confidence values of that given batch (In this example $\gamma = 0.66$) . Only the images with a confidence above this dynamic threshold contribute to the unsupervised loss.}
    \label{fig:OOD-diagram}
\end{figure}

To combat the effects of labeled and unlabeled samples coming from different distributions on current SSL methods (see results of Mean Teacher and MixMatch in \cref{table:oodresults}), we introduce out-of-distribution masking. The goal of out-of-distribution masking is to mask out the unlabeled samples that the classifier has the least confidence in, computing gradients only on samples that have a confidence above a \textit{moving threshold} and are thus likely in-distribution samples (see \cref{fig:OOD-diagram}).

It is important that the threshold for masking samples is not static, as over the course of training, we found that entropy minimization tended to force confidence values on most unlabeled samples above a specified static threshold and render the threshold useless. To find a dynamic threshold for each training step, we specify a hyperparameter $0 \leq \gamma \leq 1$ that dictates what percentage of unlabeled samples to mask. We then exclude samples that have confidence values in the bottom $\gamma * 100$\% from training. Intuitively, $\gamma$ can be thought of as the level of "noise" present in the unlabeled dataset.

\begin{figure}[t!]
    \includegraphics[width=1.\linewidth]{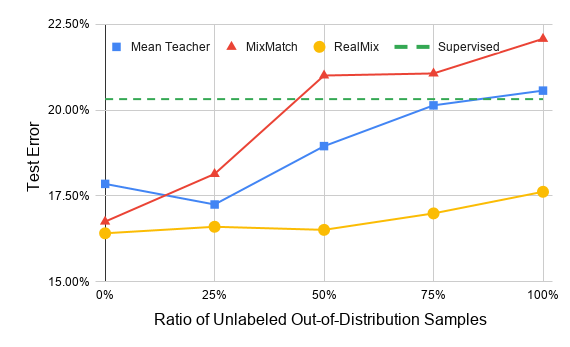}
    \caption{Error rate comparison of RealMix to other state of the art methods on the distribution mismatch experiment. All of the experiments are run using 6 animal classes from CIFAR10 with 400 samples per class as labeled data, and varying the overlap of animal classes that make up the unlabeled data. For example, at 0\% mismatch the unlabeled distribution is made up of 4 animal classes and at 100\% mismatch the unlabeled distribution is made up of 4 non-animal classes. We present supervised baseline results using the 2400 labeled samples, which achieves an error rate of 20.32\%.}
    \label{fig:OOD}
\end{figure}

Out-of-distribution masking helps to make RealMix extremely effective at mitigating unlabeled data mismatch, as RealMix is able to maintain performance above a supervised baseline no matter the amount of induced mismatch (see \cref{table:oodresults} and \cref{fig:OOD}). We also perform an ablation on $\gamma$ in \cref{table:ablationood} to show that out-of-distribution masking boosts performance even if the optimal $\gamma$ value is not found.

\subsection{Training Signal Annealing}

Semi-supervised learning algorithms have been evaluated on labeled data set sizes as few as 250 labels while the unlabeled data collections are often orders of magnitude larger \cite{mixmatchBerthelot, realisticevalOliver, ictVerma, udaXie}. To mitigate the effects of overfitting to such small quantities of labeled data samples, Xie et al.~\cite{udaXie} introduces training signal annealing (TSA). TSA delays the release of training signal based on a training schedule (logistic, linear, exponential) to limit training on labeled samples that the classifier is already confident about. We find TSA to help with training on 250 labeled samples or less using a linear schedule. 

\section{Experiments}
\label{experiments:main}
In the following sections, we show RealMix's performance on benchmark datasets, a distribution mismatch experiment, comparison to transfer learning, and an ablation study on its components.

To allow for comparison with prior SSL techniques, we follow the WRN-28-2 architecture~\cite{wrnZagoruyko}, hyperparameter selection (for $\alpha$ and $\lambda$), and evaluation procedure described by Berthelot at al.~\cite{mixmatchBerthelot} (which uses weight decay and an exponential moving average of model parameters). A key difference is that we train only for 500k iterations and use only 1 GPU, similarly to Oliver et al.~\cite{realisticevalOliver} to emulate a more realistic training setup. We report uncertainty values according to the standard deviation across 2 random seeds where possible. We also base our code implementation of RealMix and other SSL methods presented in this paper off of those created by Berthelot at al.~\cite{mixmatchBerthelot} in order to provide the research community with reproducible results.

\subsection{Baselines}
We report baseline results for the CIFAR10 and SVHN experiments from the $\Pi$-Model~\cite{temporalensemblingLaine}, VAT~\cite{vatMiyato}, Mean Teacher~\cite{meanteacherTarvainen} from those presented in ~\cite{mixmatchBerthelot}, and re-run MixMatch~\cite{mixmatchBerthelot} and RealMix according the settings described in the previous section. For the distribution mismatch experiment (illustrated in \cref{fig:OOD}), we additionally re-run and report results for Mean Teacher.
\subsection{Results}

\subsubsection{CIFAR10 and SVHN}

\begin{figure}[t!]
    \includegraphics[width=1.\linewidth]{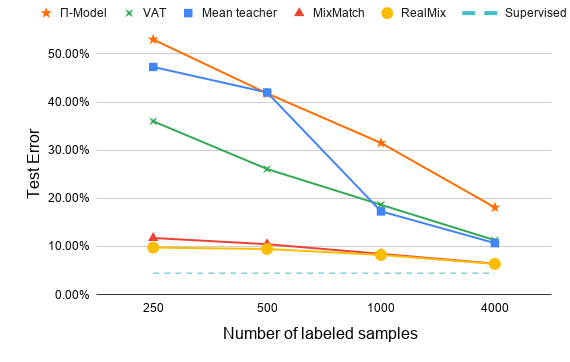}
    \caption{Results of SSL algorithms on CIFAR10 across varying labeled set sizes. Note that results $\Pi$-Model, VAT, Mean Teacher come from Berthelot et al.~\cite{mixmatchBerthelot} which are run for 500k iterations more than RealMix and MixMatch experiments. RealMix achieves state-of-the-art performance on CIFAR10 with 250 labels with an error rate of $9.79$\%, while the supervised baseline trained on all 50000 CIFAR10 samples achieves error of 4.48\%.}
    \label{fig:CIFAR10-results}
\end{figure}

We compare RealMix and prior SSL methods on the benchmark datasets CIFAR10 and SVHN, with results visible in \cref{table:cifar10}, \cref{fig:CIFAR10-results} and \cref{table:svhn}. The typical evaluation method for SSL methods is to discard all but a number of labels, reporting performance across varying labeled set sizes. For CIFAR10, we evaluate RealMix and MixMatch \cite{mixmatchBerthelot} on 4 labeled set sizes (250, 500, 1000, 4000) and present the results found by Berthelot et al.~\cite{mixmatchBerthelot} for $\Pi$-Model, VAT, Mean Teacher. For SVHN, we evaluate RealMix and MixMatch on 2 labeled set sizes (250, 4000) and compare them with results found by Berthelot et al.\cite{mixmatchBerthelot} for $\Pi$-Model, VAT, Mean Teacher. Note that these 3 models are run for 500k iterations more than the RealMix and MixMatch experiments, leaving room for further improvement on RealMix given a larger training budget.

\textbf{We find that RealMix sets a new state-of-the-art on CIFAR10 with 250 labels, with an error rate of $9.79$\% and 17\% reduction in error from the current state-of-the-art MixMatch.} Compared to the fully-supervised baseline with an error rate of $4.48$\%, RealMix is able to use 200x fewer labels to capture over 94\% of the test accuracy. We also find that RealMix is competitive with MixMatch on SVHN across labeled set sizes.
\begin{center}
\begin{table}[h]
\centering
\begin{tabular}{lrr}
    \hline
      Method & 250 Labels & 4000 Labels \\ \hline
      $\Pi$-Model ~\cite{temporalensemblingLaine} & $53.02$ & $18.13$\\
      VAT ~\cite{vatMiyato} & $36.03$ & $11.32$\\
      Mean Teacher \cite{meanteacherTarvainen} & $47.32$ & $10.72$\\
      MixMatch \cite{mixmatchBerthelot} & $11.78$ & $\bf{6.45}$  \\\hline
      \textbf{RealMix}    & $\bf{9.79 \pm 0.75}$ & $\bf{6.39 \pm 0.27}$ \\
    \end{tabular}
\caption{Results comparing error of RealMix to other SSL methods on CIFAR10 with 250 and 4000 labeled samples. The supervised baseline trained on all 50000 CIFAR10 samples achieves error of 4.48\%.}
\label{table:cifar10}
\vskip -0.3in
\end{table}
\end{center}

\begin{table}[h]
\centering
\begin{tabular}{lrr}
    \hline
      Method & 250 Labels & 4000 Labels \\ \hline
      $\Pi$-Model~\cite{temporalensemblingLaine} $\dagger$ & $17.65 \pm 0.27$ & $5.57 \pm 0.14$\\
      VAT ~\cite{vatMiyato} $\dagger$ & $8.41 \pm 1.01$ & $4.20 \pm 0.15$\\
      Mean Teacher \cite{meanteacherTarvainen}$\dagger$ & $6.45 \pm 2.43$ & $3.39 \pm 0.11$\\
      \textbf{MixMatch} \cite{mixmatchBerthelot} & $\bf{3.63} \pm 0.24$ & $\bf{3.07} \pm 0.14$  \\\hline
      \textbf{RealMix} & $\bf{3.53 \pm 0.38}$ & $\bf{3.13 \pm 0.11}$ \\
\end{tabular}
\caption{Results comparing error of RealMix to other SSL methods on SVHN with 250 and 4000 labeled samples. The supervised baseline trained on all 73257 SVHN samples achieves error of 2.72\%.}
\label{table:svhn}
\end{table}

\subsubsection{Distribution Mismatch}
\label{experiments:ood}
Oliver et al.\cite{realisticevalOliver} introduced a distribution mismatch experiment using CIFAR10 to evaluate the robustness of SSL methods to out-of-distribution examples in unlabeled data. By evaluating robustness to mismatch, a practitioner can determine in which situations SSL may be preferable to using labeled samples alone.

CIFAR10 contains two sets of classes: animals (bird, cat, deer, dog, frog, horse) and transportation (airplane, automobile, ship, truck). We simulate a mismatch by making the labeled distribution consist of the 6 animal classes each with 400 labels and varying the overlap of animal classes that make up the unlabeled distribution. For example, at 0\% mismatch the unlabeled distribution consists of 4 classes that are all animals and at 100\% mismatch, the unlabeled distribution consists of the 4 transportation classes. We evaluate RealMix, MixMatch, and Mean Teacher on varying levels of mismatch (0\%, 25\%, 50\%, 75\%, 100\%) and present our results in \cref{fig:OOD}. 

Surprisingly, \textbf{RealMix is able to surpass baseline performance on the 6 animal classes alone at all levels of mismatch.} Our ablation study (results in \cref{table:ablationood}) shows that RealMix is robust to unlabeled distribution mismatch as a result of out-of-distribution masking. Both MixMatch and Mean Teacher are able to surpass baseline performance with limited mismatch, but perform far worse when more significant amounts of mismatch (75\% and 100\%) are introduced.

Notably, RealMix is able to surpass baseline perfomance even when the unlabeled classes share no overlap with labeled classes. This would suggest that the classifier is able to learn from unlabeled data that is out-of-distribution, which we hypothesize to be the result of MixUp~\cite{mixupZhang} generating new samples that are still "slightly" in-distribution. We also selected values of the hyperparameter $\gamma$ for $OODMask_{\gamma}(x)$ as $0, 0.20, 0.40, 0.60,$ and $0.85$ respectively for the levels of mismatch to represent the expected percentage of unlabeled mismatch. These $\gamma$ values were not tuned and RealMix's performance on this experiment could presumably improve further. We also hope that future work in SSL considers out-of-distribution robustness as a key evaluation, as it not always true in real-world settings that unlabeled and labeled data arise from the same distribution.

\subsubsection{Transfer Learning}
\label{transferleanring}

Transfer learning is often an attractive first option when faced with limited quantities of labeled data, which we study following the findings of Oliver et al.\cite{realisticevalOliver} that transfer learning may be a preferable alternative to SSL. We pre-trained a classifer on ILSVRC-2012~\cite{imagenet} downsampled to 32x32 and then fine-tuned it on CIFAR10 at 250 and 4000 labels.

We find that RealMix (as well as MixMatch) outperform transfer learning and finetuning on labeled data alone, even when there is overlap in the CIFAR10 and ILSVRC-2012. This suggests that the error rates of $20.60$ and $8.45$ are upper bounds on the performance using transfer learning and fine-tuning. We also find transfer learning can be \textbf{complementary} to SSL. Specifically, we set a new state-of-the-art on CIFAR10 with 250 labels and reduce the error rate to just $8.48$\%. We also attempted transfer learning on SVHN, and found that SSL methods performed far better than transfer learning - likely because the datasets are quite different. 

\begin{center}
    \begin{table}[h]
\centering
\begin{tabular}{lrr}
    \hline
      Method & 250 Labels & 4000 Labels \\ \hline
      TL \& Fine-Tuning & 20.60 & 8.45\\
      MixMatch\cite{mixmatchBerthelot} & 11.78 & 6.45\\
      RealMix & 9.78 & 6.39 \\\hline
      \textbf{RealMix + TL} & $\bf{8.48}$ & $\bf{6.05}$ \\
\end{tabular}
\caption{Results comparing error of RealMix to transfer learning (from ILSVRC-2012) on CIFAR10 with 250 and 4000 labeled samples. We find that not only are recent SSL methods and RealMix able to surpass transfer learning alone, but combining transfer learning with RealMix can further improve results.}
\vskip -0.3in
\end{table}
\end{center}

\subsubsection{Ablation}

We finally perform an ablation study on two components of the RealMix algorithm: data augmentation and out-of-distribution masking (referred to as $ Extend(x)$ and $OODMask(x)$ respectively, in \cref{dataaugmentation} and \cref{alg:realmix}). 

RealMix extends unlabeled samples using 50 copies of samples augmented with CutOut\cite{cutoutDevries}, which gives us the state-of-the-art error rate on CIFAR10 with 250 labels of $9.79\%$. Using a simpler augmentation (random translation and horizontal flips) and using fewer augmented copies both give slightly weaker results (as listed in \cref{table:ablationdataaug}), suggesting that performing targeted augmentations and making more augmented copies of unlabeled data may further improve results. 

In \cref{experiments:ood} we study the effects of distribution mismatch on RealMix and claim that this is due to our use of out-of-distribution masking. In fact, \cref{table:ablationood} shows that RealMix's ability to meet or decrease the baseline error rate of $20.32$\% is indeed linked to out-of-distribution masking, and without it, error increases markedly.
\begin{center}
    \begin{table}[h]
\centering
\begin{tabular}{lrr}
    \hline
      Method & CIFAR10 on 250 Labels \\ \hline
      \textbf{RealMix} &\textbf{ 9.79} \\
      RealMix w/ Simple Aug & 10.42 \\
      RealMix w/ 25 Augs & 10.80 \\\hline\\
\end{tabular}
\caption{Results from ablation experiments on the augmentation type and amount from CIFAR10 on 250 labels. RealMix uses CutOut\cite{cutoutDevries} to generate 50 copies of unlabeled data.}
\label{table:ablationdataaug}
\vskip -0.4in
\end{table}
\end{center}

\begin{center}
    \begin{table}[h]
\centering
\begin{tabular}{lrr}
    \hline
      Method & OOD w/ 75\% Mismatch \\ \hline
      \textbf{RealMix }($\gamma = 0.60$) & \textbf{16.99} \\
      RealMix ($\gamma = 0.3$) & 20.73\\
      RealMix w/o OODMask & 22.70 \\\hline
\end{tabular}
\caption{Results from ablation experiments on out-of-distribution masking on the experiment from \cref{table:oodresults} with 75\% mismatch. Using OODMask, RealMix meets or surpasses the supervised baseline performance (error of 20.32\%) at multiple values of $\gamma$.}
\label{table:ablationood}
\end{table}
\end{center}

\section{Conclusion}

In this work we presented RealMix, a novel semi-supervised learning technique to improve classification performance even under situations when there is a significant shift between the distributions of the unlabeled and the labeled data.
RealMix is, to the best of our knowledge, the only SSL approach that is able to maintain baseline performance when there is a complete mismatch in the labeled and unlabeled distributions. This is a particularly important contribution when considering the applicability of semi-supervised learning outside of academic settings where data is scarce and often noisy.

We demonstrated that RealMix achieves state-of-the-art performance on common semi-supervised learning benchmarks such as CIFAR10 and SVHN, notably achieving an error rate of 9.79\% on CIFAR10 using 250 labels.

Additionally, we showed that using transfer learning techniques compliments our method to further reduce the error on CIFAR10 with 250 labels to just 8.48\%.

We hope that these results illustrate the practicality of semi-supervised learning in real world settings, and alongside the provided source code, will foster future research to further advance semi-supervised learning techniques.

{\small
\bibliographystyle{ieee_fullname}
\bibliography{egbib}
}

\end{document}